\documentclass[letterpaper]{article} 
\usepackage{aaai23}  
\usepackage{times}  
\usepackage{helvet}  
\usepackage{courier}  
\usepackage[hyphens]{url}  
\usepackage{graphicx} 
\usepackage{natbib}  
\usepackage{caption} 
\usepackage{amsmath}
\usepackage{algorithm}
\usepackage{algorithmic}

\usepackage{newfloat}
\usepackage{listings}

\usepackage{amsmath} 
\usepackage{xcolor}

\DeclareCaptionStyle{ruled}{labelfont=normalfont,labelsep=colon,strut=off} 
\floatstyle{ruled}
\newfloat{listing}{tb}{lst}{}
\floatname{listing}{Listing}
\title{Practical Bias Mitigation through Proxy Sensitive Attribute Label Generation}

\author {
    Bhushan Chaudhari,
    Anubha Pandey, 
    Deepak Bhatt, 
    Darshika Tiwari
}
\affiliations {
    AI Garage, Mastercard\\
    Gurgaon, India\\
    
    bhushan.chaudhari@mastercard.com, anubha.pandey@mastercard.com, deepak.bhatt@mastercard.com, darshika.tiwari@mastercard.com
}

\begin{document}

\maketitle

\begin{abstract}
Addressing bias in the trained machine learning system often requires access to sensitive attributes. In practice, these attributes are not available either due to legal and policy regulations or data unavailability for a given demographic. Existing bias mitigation algorithms are limited in their applicability to real-world scenarios as they require access to sensitive attributes to achieve fairness. In this research work, we aim to address this bottleneck through our proposed unsupervised proxy-sensitive attribute label generation technique. Towards this end, we propose a two-stage approach of unsupervised embedding generation followed by clustering to obtain proxy-sensitive labels. The efficacy of our work relies on the assumption that bias propagates through non-sensitive attributes that are correlated to the sensitive attributes and, when mapped to the high dimensional latent space, produces clusters of different demographic groups that exist in the data. Experimental results demonstrate that bias mitigation using existing algorithms such as Fair Mixup and Adversarial Debiasing yields comparable results on derived proxy labels when compared against using true sensitive attributes.


\end{abstract}
\section{Introduction}

Machine Learning has attained high success rates in practically every field, including healthcare, finance, and education, based on the accuracy and efficiency of the model's outcome \cite{AIApplication1, AIApplication2}. However, these models are biased and exhibit a propensity to favor one demographic group over another in various applications, including credit and loan approval, criminal justice, and resume-based candidate shortlisting \cite{BiasAI1, BiasAI2, BiasAI3}. The idea of fairness has received a lot of attention recently to combat the discrimination from the outcome of ML models \cite{FairnessAI1, FairnessAI2, FairnessAI3}.

The existing bias mitigation techniques \cite{FairnessAI4, FairnessAI5, FairnessAI6} can be classified into three categories: pre-processing  \cite{FairnessAI4, PreProcessingFairnessAI1}, post-processing \cite{FairnessAI3, PostProcessingFairnessAI1} and in-processing \cite{FairnessAI6, FairnessAI1}. While pre-processing bias mitigation techniques attempt to transform the input before feeding it to the model for training, post-processing strategies filter out the output through certain transformations. In order to produce fair output, in-processing strategies strive to learn bias-invariant models by imposing certain constraints during training. Nevertheless, most state-of-the-art algorithms require information about sensitive attributes to produce an unbiased model. However, in practice, these sensitive attributes are inaccessible due to difficulties in data collection, privacy, and legal constraints imposed by the government, like General Data Protection Regulation(GDPR) introduced by the European Union in May 2018 and Equal Credit Opportunity Act \cite{FairnessWOSensitive1, FairnessWOSensitive2}. 


Fairness is challenging to achieve in the absence of sensitive attributes due to a lack of supervision. While sensitive attributes are inaccessible in the real-world setting, it has been found that some non-sensitive attributes have strong correlations with the sensitive features, which leads to bias propagating through AI models\cite{BiasPropagationWOSentiveLabels2, BiasPropagationWOSentiveLabels1}. For instance, Hispanic and black populations have a higher proportion of younger people, resulting in the correlation between age and race \cite{ExampleCorrelation}. Similarly, zip codes can be correlated with race. Hence, the bias gets embedded in the non-sensitive attributes that are used in the model training. Based on this hypothesis, a few initial efforts have been made to mitigate bias in the absence of protected attributes \cite{FairnessWOSensitive1, FairnessWOSensitive2, FairnessWOSensitive3, FairnessWOSensitive4}. The most recent approach \cite{FairnessWOSensitive4} identifies related features that are correlated with the sensitive attributes and would further minimize the correlation between the related features and the model’s prediction to learn a fair classifier with respect to the sensitive attribute. However, identification of related features require domain knowledge and access to sensitive attributes to determine the correlation.

This research aims to provide proxy labels for sensitive attributes to make the present bias mitigation approaches suitable for real-world applications where access to protected attributes during model training is constrained. Ideally the likelihood of a positive outcome should be the same regardless of person's protected group. However in real life this does not hold true. The group which has more likelihood of getting positive outcome just because of their protected attribute is referred as favourable group  and group which has more likelihood of getting negative outcome just because of their protected attribute is referred as unfavourable group in this paper. We determine proxy for favorable and unfavorable groups by leveraging the bias information embedded in the non-sensitive features available in the given dataset. This proxy sensitive labels can then be passed as an input into the existing bias mitigation techniques. Thus we address the bottleneck in the applicability of the existing bias mitigation to real-world applications. We have proposed a novel pipeline that involves two stages: (1) Stage-1: Learn embeddings using self-supervised learning that captures inter-feature relationships and, consequently, latent bias information. (2) Stage-2: Generate proxy for demographic groups by clustering the samples based on the embeddings obtained from Stage-1. Further, experimental analysis reveals that identical results can be observed by using the proxy labels in the current bias mitigation technique as opposed to the genuine labels of sensitive qualities.



\section{Related Work}

A substantial amount of work has been done to address and mitigate bias in data sets and models\cite{BiasAI1, BiasAI3, ourPaper1}. Based on the point of intervention of the modeling stage, bias mitigation techniques broadly fall into three categories: pre-processing, in-processing, and post-processing. Pre-processing techniques underpin the first stage of the modeling and transform the training data so that the underlying discrimination is removed  \cite{PreProcessing, pre-processing1, pre-processing2, pre-processing3}. These techniques reduce or eliminate the correlation between sensitive attributes and other features, including the target labels. Unfortunately, due to the blindness of these techniques to model's inference of the data, some level of bias still can creep into the model predictions. In-processing techniques modify learning algorithms to remove bias during the model training process. Most of the algorithms in this category solve constraint optimization problem for different fairness objectives. To ensure independence between predictions and sensitive attributes, \cite{inprocessing1} regularizes the covariance between them. \cite{FairnessAI6} minimizes the disparity between the sensitive groups by regularizing the decision boundary of the classifier. \cite{adverserial_debiasing} proposed a data augmentation strategy for optimizing group fairness constraints such as equalized odds and  demographic parity. Another efficient algorithm \cite{Fair_mixup}, tries to maximize the predictor's ability to predict the ground truth while minimizing the adversary’s ability to predict the sensitive attribute. Post-processing techniques treat the learned model as a black-box model and try to mitigate bias from the prediction \cite{post-processing1, FairnessAI3, PostProcessingFairnessAI1, post-processing2}. Typically, post-processing algorithms select a subset of samples and adjust the predicted labels accordingly. An intriguing finding is that any sample can be altered to meet the requirements of group fairness because the metrics are expectations. The papers \cite{FairnessAI3, PostProcessingFairnessAI1} choose samples at random, whereas \cite{post-processing1} choose the samples with the greatest degree of uncertainty, reflecting the human tendency to give unprivileged groups the benefit of the doubt.



Most of the current algorithms have restrictions on their use in real-world scenarios since they need access to protected attributes for bias mitigation. Very recently efforts have been made towards bias mitigation in the absence of sensitive attributes \cite{FairnessWOSensitive1, FairnessWOSensitive2, FairnessWOSensitive3, FairnessWOSensitive4}. \cite{FairnessWOSensitive1} introduced a framework based on bayesian variational autoencoders that relies on knowledge of  causal graph to derive proxy. The algorithm estimates proxy in a multi dimensional space and then uses this generated proxy to remove bias from the model. But, since the proxy are generated in a multi dimensional space, they cannot be generalised to other bias mitigation algorithms. The paper \cite{Fairness_via_representation} introduced a framework wherein it only performs debiasing on the classification head. The algorithm neutralizes the training samples that have the same ground truth label but with different sensitive attribute annotations. Proxy generation for the sensitive attributes is done by training a bias intensified model and then annotating samples based on its confidence level. However, the algorithm makes a strong assumption that bias-amplified model tends to assign the privileged group more desired outcome whereas assigning the under-privileged group a less-desired outcome based on the obtained prediction scores. The most recent approach \cite{FairnessWOSensitive4} identifies related features that are correlated with the sensitive attributes and would further minimize the correlation between the related features and the model’s prediction to learn a fair classifier with respect to the sensitive attribute. To identify the related features, however, this method needs access to sensitive attributes to determine the correlation.


\begin{figure*}
    \centering
    \includegraphics[width=\textwidth]{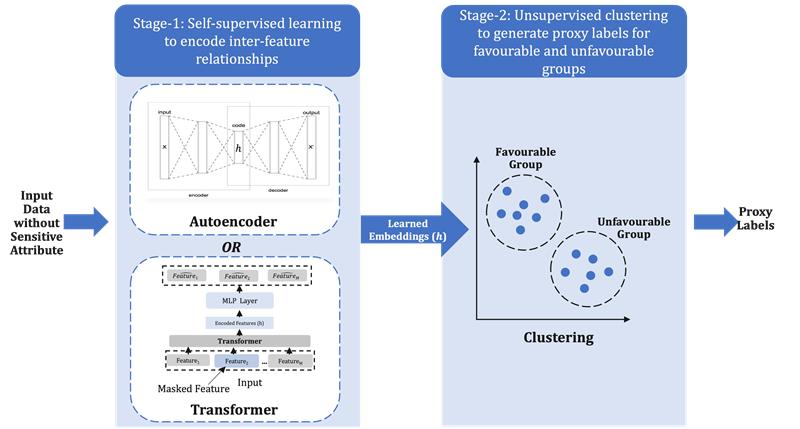}
    \caption{Proposed pipeline for proxy sensitive attribute label generation.}
    \label{fig:pipeline}
\end{figure*}

\section{Methodology}

It is widely established that bias propagates to the models even when protected attributes are not used during training \cite{FairnessWOSensitive1, FairnessWOSensitive2, FairnessWOSensitive3, FairnessWOSensitive4}. This is attributed to the frequent incorporation of protected attribute data into other correlated non-protected attributes. Zip codes, for instance, can be associated to the race attribute. Based on this hypothesis, we utilize the non-protected attributes to obtain proxy-sensitive labels. Assuming the availability of all variables except the protected attribute, our goal is to recover all the latent information associated with the protected attribute embedded into the available non-protected features.

This section outlines our suggested method for generating a proxy for a sensitive protected attribute. We break the objective down into two stages. In the first stage, we utilize self-supervised learning to produce the contextual embedding of the input samples. Our goal is to learn an embedding with maximum information about the protected attribute. In the second stage, we obtain proxy labels for favorable and unfavorable groups using an unsupervised clustering approach on the embedding obtained from the first stage. Finally, we pass the generated proxy through existing state-of-the-art bias mitigation algorithms to mitigate bias from any model. Figure \ref{fig:pipeline} outlines the proxy-generation pipeline.

\begin{figure*}
    \centering
    \includegraphics[width=\textwidth]{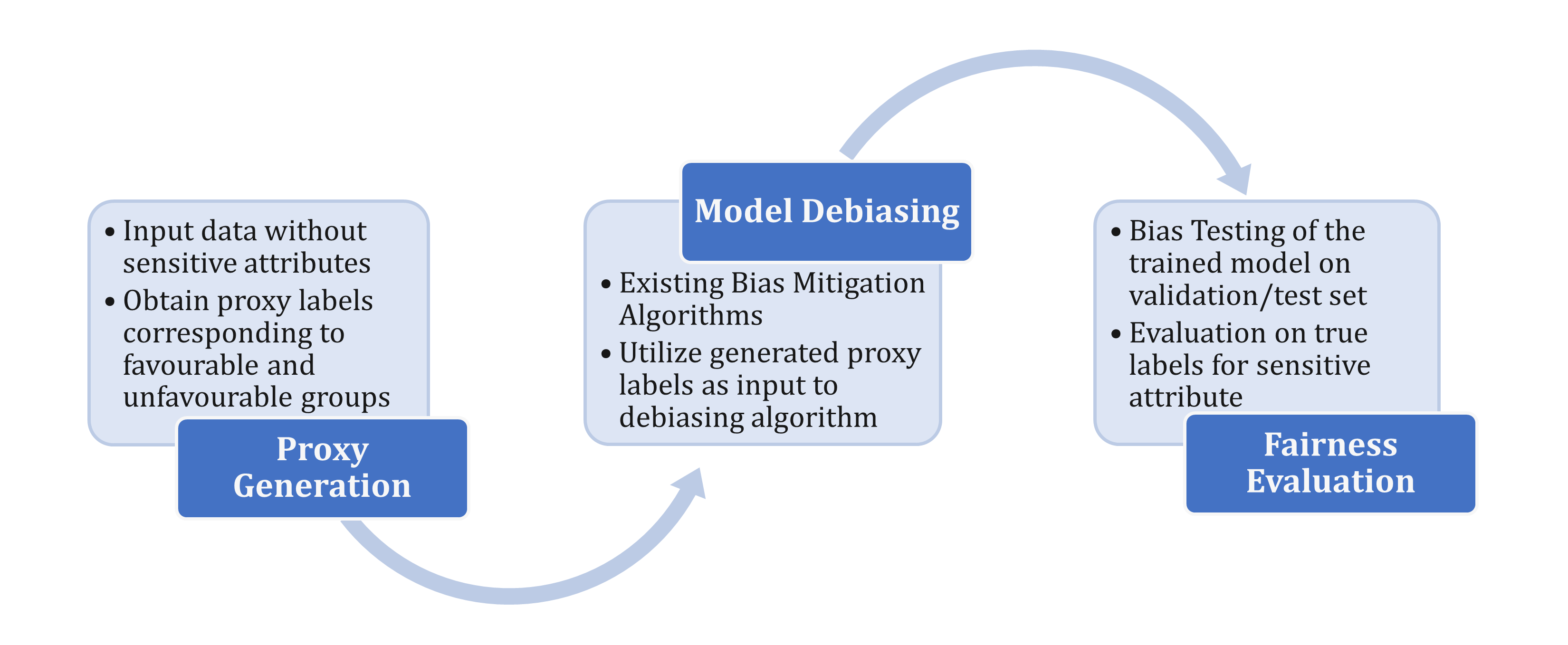}
    \caption{Overall pipeline for bias mitigation and evaluation}
    \label{fig:proxy_eval}
\end{figure*}

\subsection{Proxy Generation for Sensitive Attribute}

\textbf{Stage-1:} In the first stage, as shown the figure \ref{fig:pipeline}, we obtain contextual embedding of the input samples. Towards this goal, we train neural network architectures in a self-supervised fashion to efficiently encode inter-feature relationships. In this paper, we have experimented with two neural network architectures: (1) Auto-encoders and, (2) Transformers. 

We train an auto-encoder on the reconstruction task to obtain embeddings containing crucial input data details. An auto-encoder consists of encoder and decoder modules. In the encoding operation, we pass the input feature vector that gets mapped to a lower dimensional latent representation. In the decoding operation, the original input data gets reconstructed back from the latent representation. We trained the network on a reconstruction loss that minimizes the mean square error between the input and output embeddings. Input data $X$ is passed through the encoder to get latent representation $h$ and then reconstructed as $\hat{X}$ by the decoder as shown in the equations \ref{AE_hidden} and \ref{AE_reconstruction}. We train the network on reconstruction loss $Loss_{AE}$ as shown in equation \ref{AE_recon_loss} where $n$ represents the number of data points in a batch. Here, $f1$ and $f2$ are activation functions, $W$ is weight matrix and $b$ is bias.The latent embeddings obtained from the encoder module contain information about the protected attribute as it is generated from features that are correlated with the protected attribute.

\begin{equation}
h = f1(W_{i}*X + b_i)
\label{AE_hidden}
\end{equation} \\
\begin{equation}
\hat{X} = f2(W_{j}*h + b_j)
\label{AE_reconstruction}
\end{equation} \\
\begin{equation}
Loss_{AE} = \frac{1}{n}\Sigma_{i=1}^{n}|{X_i -\hat{X_i}}|
\label{AE_recon_loss}
\end{equation} \\


We experimented with another neural network architecture called Transformer with a similar goal. Transformers utilize a self-attention \cite{Attention} mechanism to learn the embeddings. To compute self-attention, first, three vectors, Query(Q), Key(K), and (V), are learned corresponding to each feature in the input, and then the attention is computed as shown in the equation \ref{attention}. Finally, self-attained embeddings $h$ are obtained as shown in the equation \ref{Emb}. We train the Transformer on a self-supervised learning task called Masked Language Modelling (MLM). Towards this, 15$\%$ of the input data fields are chosen randomly and replaced with a masked token. The Transformer then processes samples to produce contextual row embeddings. The MLM head, made up of MLP layers, reconstructs the original fields from these row embeddings. The model is trained end-to-end by minimizing cross-entropy loss as shown in the equation \ref{T_Loss}. The loss is calculated only on masked fields. The latent embeddings ($h$) obtained from the transformer contain information about the protected attribute due to its inherent property to learn the inter-feature relationships.


\begin{equation}
Attention(Q, K, V) = softmax(\frac{QK^T}{\sqrt{d_k}})V
\label{attention}
\end{equation} 

\begin{equation}
head_i = Attention(Q W^Q_i, K W^K_i, V W^V_i)
\label{MultiAttention}
\end{equation} 

\begin{equation}
h = Concat(head_1, ..., head_h)W^O
\label{Emb}
\end{equation}


\begin{equation}
p_{i} = Softmax(MLP(h))
\label{probability}
\end{equation} \\
\begin{equation}
Loss_{T} = -\sum_{c=1}^My_{i}\log(p_{i})
\label{T_Loss}
\end{equation} \\


Further to ensure that the generated embeddings do not corresponds to the true labels of the downstream classification task, we have trained the above described neural network models on KL Divergence loss. KL divergence loss historically has been used in classification tasks to ensure class separation between two different labels. The KL divergence loss is based on the information theoretic measure of the Kullback-Leibler (KL) Divergence, which measures the difference between two probability distributions. By introducing the KL divergence loss, the model is able to learn the distinction between the two different labels better, thus leading to improved embedding generation which contains information related to protected attribute and not downstream task labels.

In order to implement the Kullback-Leibler (KL) divergence in the proposed neural network architecture, a multi-layer perceptron (MLP) layer has been applied on the generated embedding vectors. In the autoencoder, the MLP is applied on top of the latent vectors, while in the transformer, the MLP is fed with the contextual vector (h). The calculation of the KL loss on top of the MLP depends on the input embedding vector. Specifically, the input embedding vector is fed into the MLP, which will generate the probability distribution. Then, the KL divergence between the probability distribution and the target distribution is calculated. This KL loss is then used to optimize the MLP weights and biases.\\

\textbf{Stage-2:} In the second stage, as shown the figure \ref{fig:pipeline}, we use an unsupervised clustering algorithm to identify various groups in the embeddings obtained from the previous stage. As we know, clustering is a subjective statistical analysis, and there are many algorithms suitable for each data set and problem type. In this paper, we have experimented with centroid-based and hierarchical clustering algorithms. In particular, we have experimented with K means, Hierarchical and BIRCH to obtain two clusters that serves as a proxy for favourable and unfavourable groups. We further evaluate the performance of generated proxy from each clustering algorithm on bias mitigation.

\subsection{Bias Mitigation Through Generated Proxy Sensitive Attribute}
Once the proxy labels are obtained corresponding to the favourable and unfavourable groups, we pass them as input to the existing bias mitigation algorithms. In this paper, we have experimented with two widely used benchmarks for bias mitigation: Adversarial Debiasing and Fair Mixup. Both algorithms require labels corresponding to the protected attribute in the input. We pass the proxy for the protected attribute obtained from the proposed pipeline as input to de-bias the model. We have compared the performance on bias mitigation with the true labels and proxy labels for the protected attributes in the results section. However, for fairness evaluation, we use true sensitive labels. Figure \ref{fig:proxy_eval} shows the pipeline for bias mitigation and fairness evaluation. 


\section{Experimental Details}

\subsection{Dataset Description}
We have evaluated the proposed pipeline on the Adult Income Dataset, generated from 1994 US Census. The objective of the dataset is to predict the income level based on personal individual information. The target variable,Y takes a binary value depicting salary $\leq$ 50K or salary $>$ 50k.The dataset consist of 14 independent attributes and the field 'Gender' is considered as a sensitive attribute in our case. It takes up two values, namely 'Male' and 'Female'.The dataset is imbalanced: only
24\% of the samples belong to class 1, out of which only 15.13\% are females. The dataset consist of 48,842 independent rows. During the training of our model, we do not take into account the information provided by the 'Gender' attribute.\\

\begin{table*}
\begin{center}
\begin{tabular}{|c| c| c| c|}
    \hline
    \multicolumn{1}{|c|}{\textbf{Bias Mitigation Algortihm}} & \multicolumn{1}{|c|}{\textbf{Average Precision}} &
    \multicolumn{1}{|c|}{
    \textbf{SPD}} &
    \multicolumn{1}{|c|}{
    \textbf{EOD}}\\
    \hline\hline 
     w/o Bias Mitigation & \textbf{0.8} & 0.2 & 0.11 \\
     \hline
     Fair Mixup & 0.78 & \textbf{0.1} & \textbf{0.03} \\
    \hline
    \text{Adversarial Debiasing} & 0.78 & 0.19 & 0.05\\
    \hline
\end{tabular}
\caption{Performance evaluation of existing bias mitigation algorithms.}
\label{tab:fairness-true}
\end{center}
\end{table*}

\subsection{Implementation Details}
 
We have implemented the proposed pipeline in the Pytorch framework. All the experiments were performed on Ubuntu 16.04.7 with the Nvidia GeForce GTX 1080Ti GPU. 16GB of RAM was utilized while experimenting on the Adult Income Dataset.

In Stage-1, we experimented with two embedding generator networks, autoencoders and transformers. The autoencoders used in the algorithm consist of one hidden layer. The hidden layer's output receives ReLU activation while its input receives Tanh activation. The model was trained for 200 epochs with a batch size of 32 and a learning rate of 0.001 using Adam as the optimizer. The Transformer architecture contains only the encoder module. Three encoder blocks are used with six attention heads. Each encoder module is a feed-forward network with 128 hidden units. We used the implementation of the Transformer provided in the hugging face library. In Stage-2, we experimented with K-Means, BIRCH, and Hierarchical clustering algorithms to generate proxy labels for protected attributes. We utilize the implementation of these clustering algorithms given in python's sklearn library. 

We utilize all the data samples to train our proposed pipeline to obtain the proxy labels for sensitive attributes. Next, we randomly split the dataset into 80-20 train and test split and train the classification model using bias mitigation algorithms on the train set. We employ existing bias mitigation algorithms like Adversarial debiasing\cite{adverserial_debiasing} provided in the IBM AIF360 toolkit and fair-mixup\cite{Fair_mixup} an open-source solution that is accessible on GitHub. During training, we use the generated proxy instead of the actual labels of the protected attribute and assess performance on the protected attribute's actual labels.


\subsection{Fairness Metrics}

Fairness in machine learning measures the degree of disparate treatment for different groups (e.g., female vs. male), or individual fairness, emphasizing similar individuals should be treated similarly. There exists various metrics in the literature to quantify fairness, each focusing on different aspects of fairness. We have used two popularly used metrics: Statistical Parity Difference (SPD) and Equalized Odds Difference (EOD). \\

\textbf{Statistical Parity Difference (SPD) } : A classifier is considered fair if the prediction Y on input features X is independent from the protected attribute S. The underlying idea is that each  demographic group has the same chance for a positive outcome. \cite{FairnessMetric} 
\begin{equation}
SPD = |{P(\hat{Y}=1|S = 0)} - {P(\hat{Y} = 1|S = 1)}|
\label{SPD}
\end{equation} \\

\textbf{Equalized Odds Difference (EOD)} : An algorithm is considered fair if across both privileged and unprivileged groups, the predictor Y has equal false positive rate(FPR) and false negative rate(FNR). This constraint enforces that accuracy is equally high in all demographics since the rate of positive and negative classification is equal across the groups.The notion of fairness here is that chances of being correctly or incorrectly classified positive should be equal for every group.

\begin{equation}
\begin{aligned}
\triangle{FPR} = |\{P(\hat{Y}=1|S = 1, Y = 0) - \\ P(\hat{Y}=1|S = 0, Y = 0)\}|
\end{aligned}
\end{equation}

\begin{equation}
\begin{aligned}
\triangle{FNR} = |\{P(\hat{Y}=0|S = 1, Y = 1) - \\P(\hat{Y}=0|S = 0, Y = 1)\}|
\end{aligned}
\end{equation}


\begin{equation}
EOD = \frac{\triangle{FPR} + \triangle{FNR}}{2} 
\label{EOD}
\end{equation}\\


\vspace{-10pt}

\begin{table*}
\begin{center}
\begin{tabular}{|c| c| c| c| c|c|c|c|}
    \hline
    \multicolumn{2}{|c|}{} & \multicolumn{3}{|c|}{\textbf{FairMixup}} &
    \multicolumn{3}{|c|}{
    \textbf{Adversarial Debiasing}}\\
    \hline\hline 
    Embedding & Clustering & Avg Precision & SPD & EOD & Avg Precision & SPD & EOD\\ [0.5ex] 
    \hline
      & K-Means & 0.76 & 0.16 & 0.09  & 0.78 & 0.06 & 0.06\\
    AutoEncoder  & \text{Hierarchical} & 0.78 & 0.11 & 0.06 & \textbf{0.76} & \textbf{0.05} & \textbf{0.01}  \\
      & \text{BIRCH} & 0.77 & 0.13 & 0.07 & 0.79 & 0.16 & 0.09 \\
     \hline
     
       & K-Means & 0.75 & 0.09 & 0.07 & 0.79 & 0.15 & 0.04  \\
    Transformer  & \text{Hierarchical} & \textbf{0.77} & \textbf{0.07} & \textbf{0.05} & 0.77 & 0.12 & 0.04 \\
      & \text{BIRCH} & 0.75 & 0.13 & 0.07 & 0.79 & 0.11 & 0.09  \\
     \hline 
\end{tabular}
\caption{Fairness and performance results on two open bias mitigation algorithms}
\label{tab:fairness-proxy}
\end{center}
\end{table*}
\vspace{-10pt}

\vspace{-20pt}

\section{Results}


In this section, we empirically assess the effectiveness of the proxy-sensitive label obtained through the proposed pipeline. Towards this end, we pass the proxy-sensitive labels through state-of-the-art bias mitigation methods like adversarial debiasing and fair mixup and evaluate the fairness and classification performance on a public dataset called UCI Adult Income. We have reported the classification performance on Average Precision and fairness on Statistical Parity Difference (SPD) and Equalized Odds Difference (EOD).

Fair mixup and adversarial debiasing bias mitigation algorithms require protected attribute information to de-bias the models. To form the baseline, we have passed true labels of the protected attribute gender through the mentioned bias mitigation algorithms. Fair mixup has a trade-off parameter between fairness and accuracy, called lambda. We set this parameter as 0.5 for SPD and 2.5 for EOD. Table \ref{tab:fairness-true} compares the classification and fairness performance of the model trained using bias mitigation algorithms like fair mixup and adversarial debiasing against a classifier trained without using any bias mitigation algorithm. From table \ref{tab:fairness-true}, we can observe that the model trained without any bias mitigation algorithm produce an average precision of 0.8 and SPD and EOD metrics as 0.2 and 0.11. 
However, with model debiasing, we can see an improvement in SPD and EOD values proving the efficacy of the bias mitigation algorithms in achieving fairness.

In this paper, we concentrate on a more practical experimental setup, where we have assumed that the protected attributes are unavailable during model training. Here, we have used a proxy generated by our pipeline as an input to the existing bias mitigation techniques discussed above rather than the true labels of the protected attribute to test the efficacy of the generated proxy in model debiasing. With proxy-sensitive labels, we aim to achieve a similar performance as the baselines as shown in Table \ref{tab:fairness-true}.

\begin{figure}
    \centering
    \includegraphics[width=0.5\textwidth]{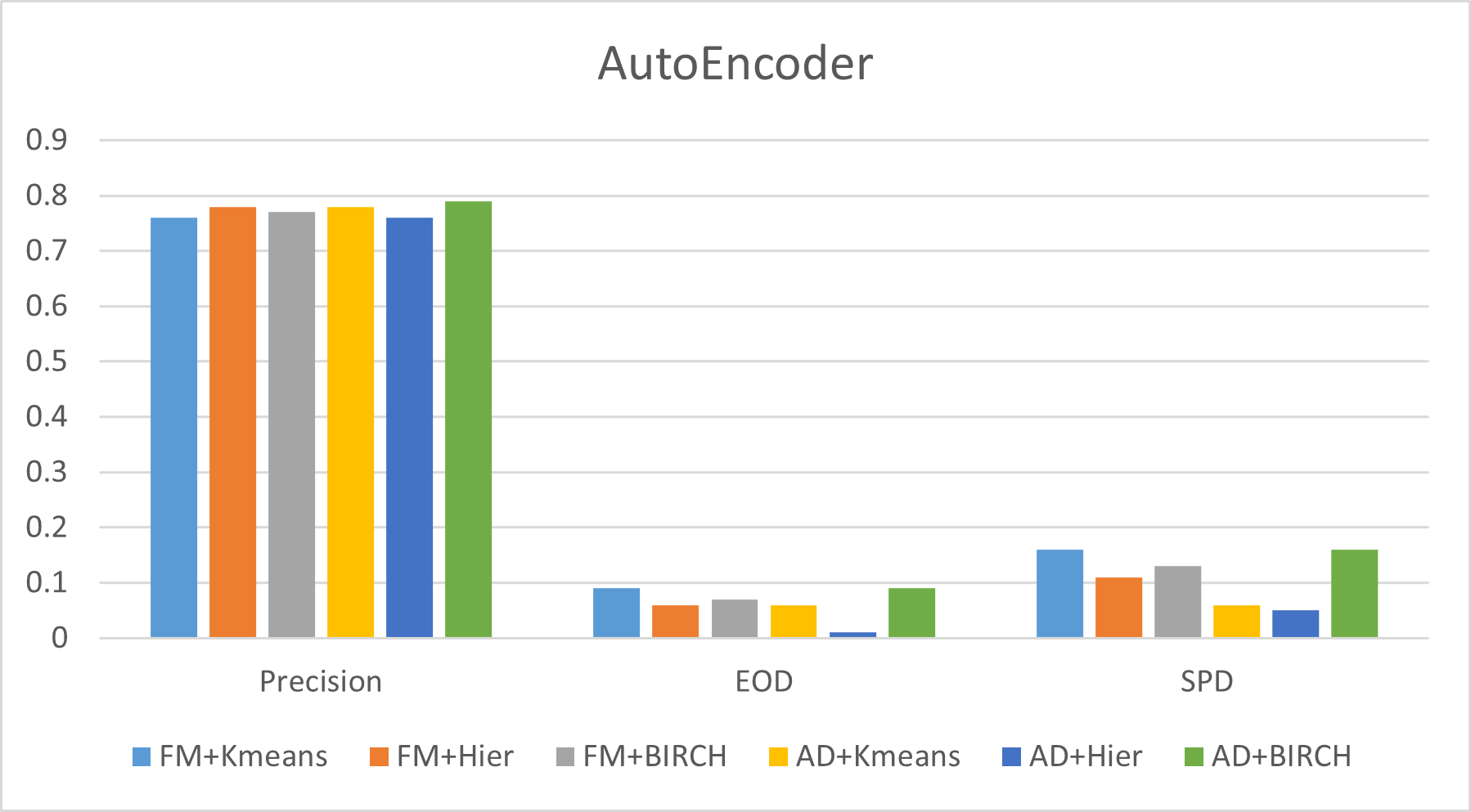}
    \caption{Average precision and fairness metrics obtained by different techniques of clustering on embeddings generated by Autoencoder architecture. Results are shown on two state-of-the-art bias mitigation techniques, Fair Mixup (FM) and Adversarial Debiasing (AD), with proxy-sensitive labels as input instead of true sensitive labels.}
    \label{fig:AE_result}
\end{figure}

\begin{figure}
    \centering
    \includegraphics[width=0.5\textwidth]{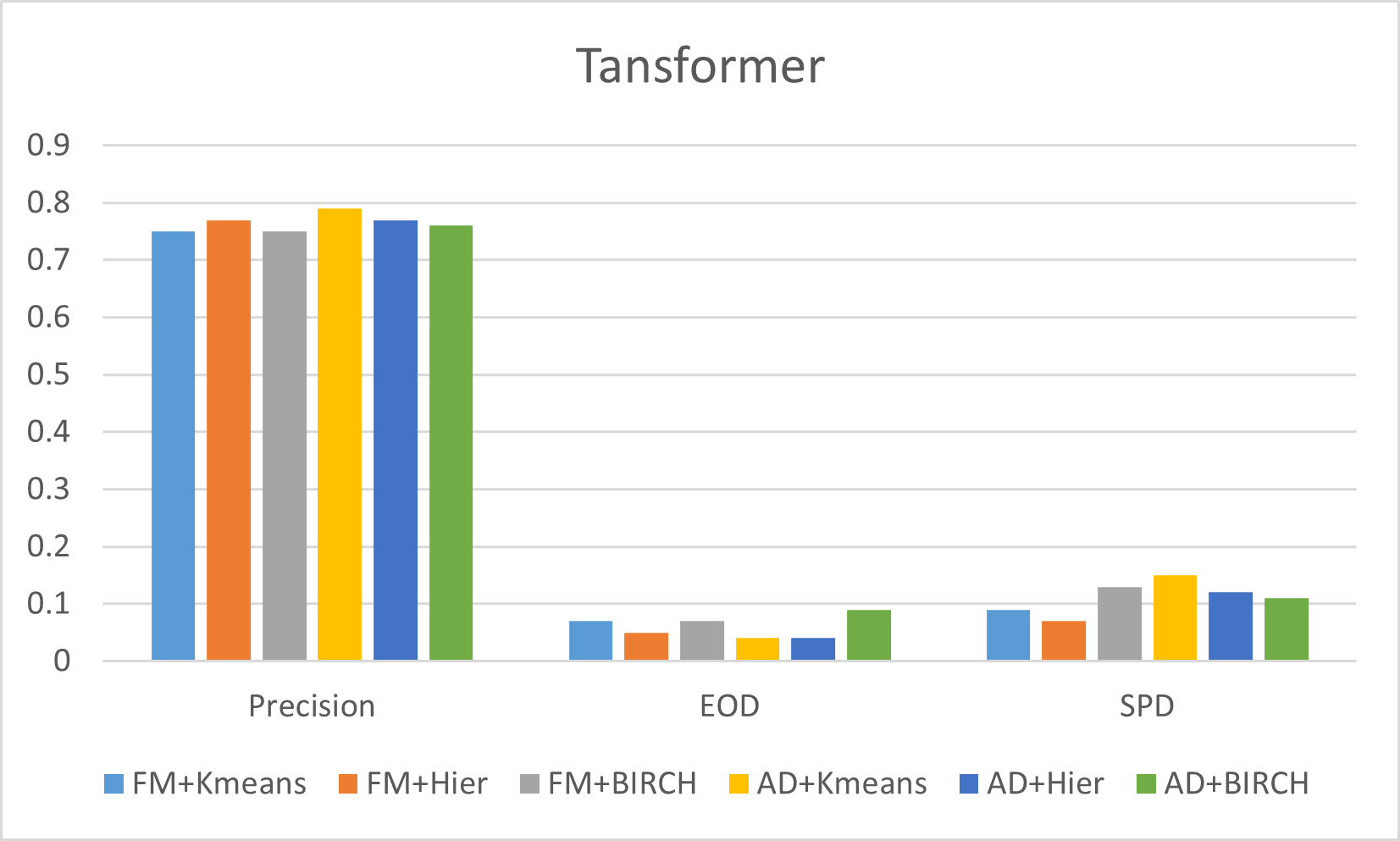}
    \caption{Average precision and fairness metrics obtained by different techniques of clustering on embeddings generated by Transformer architecture. Results are shown on two state-of-the-art bias mitigation techniques, Fair Mixup (FM) and Adversarial Debiasing (AD), with proxy-sensitive labels as input instead of true sensitive labels.}
    \label{fig:Transformer_Result}
\end{figure}

We have experimented with several algorithms in both stages to generate proxy-sensitive labels. In Stage 1, we experimented with Autoencoder and Transformer architectures to generate the embeddings. And in Stage-2, we experimented with clustering algorithms like K-means, Hierarchical, and BIRCH. Figure \ref{fig:AE_result} shows the performance of all the configurations when Autoencoder is used for embedding generation, and Figure \ref{fig:Transformer_Result} shows the performance when Transformer is utilized. From figure \ref{fig:AE_result} we can observe that the proxy generated by hierarchical clustering produces the best results with the adversarial debiasing algorithm. In this configuration, we can observe an absolute improvement of 0.14\% in SPD with comparable average precision and EOD performance when proxy labels are used instead of true labels for the sensitive attribute Gender. Figure \ref{fig:Transformer_Result} shows that with the Fair mixup algorithm, the best-performing configuration with proxy-sensitive labels has achieved an average precision of 0.77 with EOD and SPD values of 0.05 and 0.07. This performance is comparable to model performance with the true protected attribute. On the other hand, with the Adversarial debiasing algorithm, the embeddings obtained from the Transformer have led to a $1\%$ absolute lift in the average precision while improving the fairness metrics compared to the baseline model trained on true sensitive labels. 

Transformer architecture to learn embedding in the proxy generation phase produces a significant lift in fairness. The inherent properties of the transformer architecture to learn the inter-feature relationships enables it to generate informative embeddings for the tabular dataset. This is supported by the experimental results shown in figures \ref{fig:AE_result} and \ref{fig:Transformer_Result} on the Adult Income dataset. However, the choice of the modeling architectures to obtain the embedding and the clustering algorithms are dataset-dependent.

\subsection{Learned Embedding Analysis}

The performance evaluation discussed in the above section indicates that the proxy-sensitive labels can be used as a substitute for the true labels of protected attributes in the existing bias-mitigation algorithms. In this section, we analyze the quality of embeddings learned in Stage-1 of the proposed pipeline through an auxiliary prediction task similar to \cite{AuxilaryModel}.


Towards this effect, we train three linear classifiers  $C\textsubscript{Proxy}$, $C\textsubscript{True}$ and $C\textsubscript{Downstream}$ that take the embeddings as input and predict proxy attribute, true protected attribute, and target class labels respectively. Next, we compare the learned weight matrix of $C\textsubscript{Proxy}$ with $C\textsubscript{True}$ and $C\textsubscript{Downstream}$ separately using cosine similarity.

The cosine similarity between the weight vectors of $C_{Proxy}$ and $C_{True}$ is 0.25, and between $C_{Proxy}$ and $C_{Downstream}$ is 0.02. A high value of cosine similarity between weight parameters of $C_{Proxy}$ and $C_{True}$ indicates that embedding contains a substantial amount of information about the true protected attribute. In contrast, a low cosine similarity value between weights of $C_{Proxy}$ and $C_{Downstream}$ indicates that the clusters formed over the embeddings are not along the downstream prediction task.



\section{Conclusion}


Bias mitigation with no access to sensitive attributes is a challenging problem and has received little attention in the literature. Numerous relevant research studies exist on fairness in AI, but most of these studies assume that protected attributes are accessible at the time of training. This assumption limits their use in modeling scenarios where protected labels are unavailable. In an effort to reduce this dependency, we propose a novel pipeline that leverages the inherent bias information in the non-protected attributes to obtain proxy labels of protected attributes. In the current state-of-the-art bias mitigation algorithms, these proxies are passed as input rather than the true labels of the sensitive attribute. Experimental results demonstrate that model trained using generated proxy labels results in satisfactory bias metrics such as SPD and EOD with little or no reduction in detection rate. In the future, we will continue to advance our research by investigating more effective methods to incorporate additional bias information into the embedding to improve proxy labels. Additionally, we would validate the compatibility of the proposed approach with additional bias mitigation algorithms beyond the algorithms studied in this work.

\bibliography{aaai23}

\end{document}